\documentclass[journal]{IEEEtran}

\ifCLASSINFOpdf
\else
\fi

\usepackage{framed,multirow}
\usepackage{graphicx}
\usepackage{amssymb}
\usepackage{amsmath}
\usepackage{multirow}
\usepackage{booktabs}
\usepackage{array}
\newcolumntype{L}[1]{>{\raggedright\let\newline\\\arraybackslash\hspace{0pt}}m{#1}}
\newcolumntype{C}[1]{>{\centering\let\newline\\\arraybackslash\hspace{0pt}}m{#1}}
\newcolumntype{R}[1]{>{\raggedleft\let\newline\\\arraybackslash\hspace{0pt}}m{#1}}

\usepackage{lipsum}
\usepackage{changepage}
\usepackage{color,soul}
\usepackage{amssymb}
\usepackage{amsmath}
\usepackage{enumitem}   

\usepackage{float}

\usepackage{booktabs,xcolor}
\definecolor{lightgray}{gray}{0.9}

\usepackage{tikz}

\usepackage{lipsum}

\let\OLDthebibliography\thebibliography
\renewcommand\thebibliography[1]{
  \OLDthebibliography{#1}
  \setlength{\parskip}{0pt}
  \setlength{\itemsep}{0pt plus 0.3ex}
}

\usepackage{array}

\makeatletter
\newcommand{\thickhline}{%
    \noalign {\ifnum 0=`}\fi \hrule height 1pt
    \futurelet \reserved@a \@xhline
}
\newcolumntype{"}{@{\hskip\tabcolsep\vrule width 1pt\hskip\tabcolsep}}
\makeatother

\usepackage{amssymb}
\usepackage{latexsym}

\usepackage{url}
\usepackage{xcolor}

\usepackage{hyperref}

\definecolor{newcolor}{rgb}{.8,.349,.1}

\usepackage{bm}

\usepackage{pifont}
\newcommand{\cmark}{\ding{51}}
\newcommand{\xmark}{\ding{55}}

\begin{document}

\onecolumn
\begin{center}
\Huge{Improving Calibration and Out-of-Distribution Detection in Medical Image Segmentation with Convolutional Neural Networks}
\end{center}

\vspace{2mm}

\begin{center}
\normalsize
Davood Karimi, Ali Gholipour \\ Department of Radiology, Boston Children’s Hospital, Harvard Medical School, Boston, MA, USA
\end{center}

\vspace{4mm}

\begin{abstract}

Convolutional Neural Networks (CNNs) have shown to be powerful medical image segmentation models. In this study, we address some of the main unresolved issues regarding these models. Specifically, training of these models on small medical image datasets is still challenging, with many studies promoting techniques such as transfer learning. Moreover, these models are infamous for producing over-confident predictions and for failing silently when presented with out-of-distribution (OOD) data at test time. In this paper, we advocate for multi-task learning, i.e., training a single model on several different datasets, spanning several different organs of interest and different imaging modalities. We show that not only a single CNN learns to automatically recognize the context and accurately segment the organ of interest in each context, but also that such a joint model often has more accurate and better-calibrated predictions than dedicated models trained separately on each dataset. Our experiments show that multi-task learning can outperform transfer learning in medical image segmentation tasks. For detecting OOD data, we propose a method based on spectral analysis of CNN feature maps. We show that different datasets, representing different imaging modalities and/or different organs of interest, have distinct spectral signatures, which can be used to identify whether or not a test image is similar to the images used to train a model. We show that this approach is far more accurate than OOD detection based on prediction uncertainty. The methods proposed in this paper contribute significantly to improving the accuracy and reliability of CNN-based medical image segmentation models.

\end{abstract}

\footnotesize
\hspace{5mm} \textbf{Index Terms: } Medical image segmentation, convolutional neural networks, heterogeneous data, prediction uncertainty, out-of-distribution detection




\vspace{15mm}

\begin{figure*}[hbt!]
\begin{minipage}[b]{1.0\linewidth}
  \centering
  \centerline{\includegraphics[width=17.6cm]{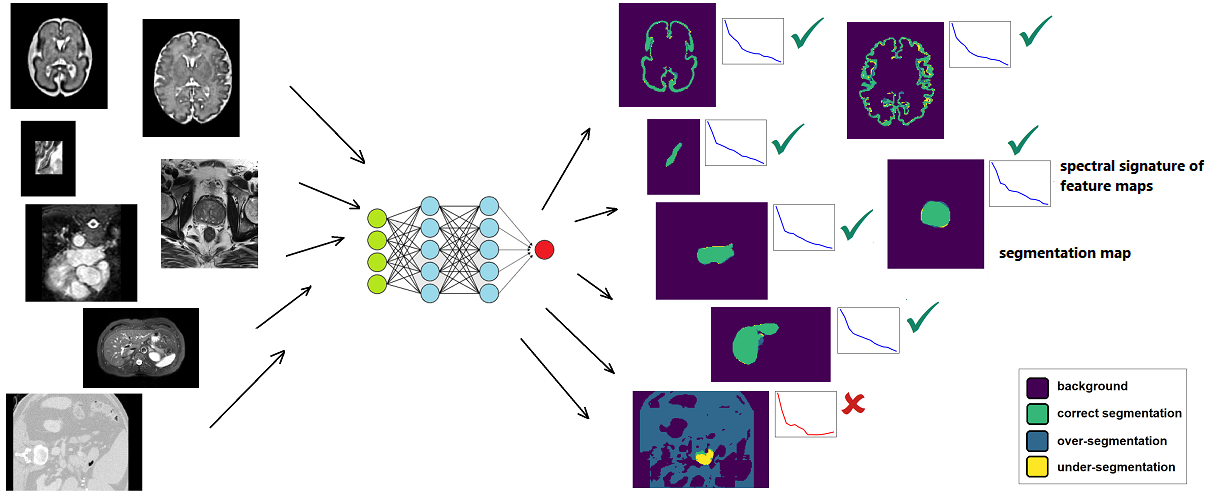}}
  \end{minipage}
\caption{We propose multi-task learning as a new approach for training CNN-based medical image segmentation models. Unlike the common approach of training a model to segment one organ in one imaging modality, we advocate for training a single model to segment different organs in different imaging modalities. We show that a standard CNN can automatically recognize the context and accurately segment different organs in different imaging modalities, without the need for any additional inputs. We show that a single model trained on a mix of heterogeneous datasets is as accurate as or even more accurate than models dedicated to individual datasets. Moreover, a model trained on heterogeneous data usually has better-calibrated predictions. For detecting out-of-distribution (OOD) data at test time, we propose a novel method based on spectral analysis of the CNN feature maps. We show that this method can detect OOD data more accurately than a method based on prediction uncertainty.}
\label{fig:methods_schematics}
\end{figure*}

\normalsize
\twocolumn

\section{Introduction}

\IEEEPARstart{M}{edical} image segmentation is an essential component of many medical image analysis and image-guided intervention pipelines. Compared with manual segmentation by an expert, computerized automatic segmentation methods have the potential to improve the speed and reproducibility of the segmentations. Classical automatic medical image segmentation techniques include such methods as region growing, level-sets, and atlas-based techniques. Recently, deep learning models, and in particular convolutional neural networks (CNNs), have shown to be excellent tools for this task.

Many recent studies have shown that CNN-based methods outperform the classical methods on various medical image segmentation tasks, often by significant margins. Because of the success of CNN-based models, various aspects of their design and training have been investigated in the past few years. Many of these studies have focused on such aspects as network architecture and loss function. However, it has been shown that factors such as more elaborate network architectures can often only marginally improve the performance of standard CNN-based medical image segmentation methods \cite{isensee2018}.

There are two main unresolved issues with regard to the application of CNNs for medical image segmentation. The first issue has to do with the training procedures and training data. Specifically, the number of manually-labeled images that are available for training is typically very small compared with many non-medical applications. This is because the number of images is small to begin with, and accurate manual annotation is costly because it depends on domain expertise. In recent years, this challenge has led to a surge of interest in such techniques as transfer learning \cite{tajbakhsh2016}, unsupervised learning \cite{cheplygina2019}, and learning from inaccurate and computer-generated annotations \cite{karimi2019noisylabel}. 

The second outstanding issue is a lack of understanding of the reliability and failure modes of these models. Deep learning models, in general, are known to produce over-confident predictions, even when the predictions are completely wrong \cite{guo2017}. In other words, there is little correlation between the confidence of a deep learning model in its predictions and how accurate the predictions actually are. Deep learning models also produce confident predictions on out-of-distribution (OOD) data, i.e., when the test data is from an entirely different distribution than the training data \cite{lee2018b,hendrycks2016}. Needless to say, there is no performance guarantee on OOD data. In fact, in general the model predictions on OOD data are not expected to be any more useful than random assignment.

In order to improve the accuracy and reliability of CNN-based medical image segmentation models for real-world clinical usage, effective solutions are needed for the above-mentioned challenges. In particular, we need methods that can train accurate and well-calibrated medical image segmentation models from limited data. Furthermore, we need methods to inform us when these models fail. The goal of this paper is to make significant contributions in addressing these challenges.

\section{Related works}

\subsection{Training procedures for CNN-based medical image segmentation models}

Large labeled datasets are considered an essential requirement for training of modern deep learning models \cite{lecun2015}. Since such datasets are difficult or impossible to come by in medical image segmentation, a range of strategies have been proposed to tackle this limitation. Here, we briefly review the most important classes of these methods.

One strategy is transfer learning \cite{pan2009b}, whereby the model is first trained on data from other domains/tasks and then fine-tuned for the intended task. Transfer learning has been reported to improve the performance of CNN-based models on many medical image segmentation tasks \cite{tajbakhsh2016,ghafoorian2017}. A limitation of transfer learning is that most of the large public image datasets include only 2D images, whereas most medical images are 3D.

An alternative to transfer learning is multi-task learning \cite{caruana1998dozen,pan2009b}. Unlike transfer learning that addresses the source and target tasks sequentially, multi-task learning aims at learning multiple tasks simultaneously. It is based on the expectation that the tasks at hand share enough similarities such that learning them simultaneously with a single model would help learn more relevant features and lead to improved generalizability. Some studies have reported successful application of multi-task learning to medical image segmentation \cite{moeskops2016deep,harouni2018universal}. However, the effect of multi-task learning on the prediction calibration of the trained model has not been explored in previous studies.

Semi-supervised and weakly-supervised methods constitute a large and diverse body of techniques \cite{chapelle2009,cheplygina2019}. In brief, these methods aim at utilizing a mix of labeled and unlabeled data or data that have not been labeled in detail. These methods have also been used in deep learning-based medical image segmentation with relative success, as in \cite{enguehard2019, baur2017}. One possibility is to use other, less accurate, automatic methods to generate approximate segmentations on large corpora of medical images and use those to train a more accurate CNN-based segmentation model \cite{ghafoorian2018, zhang2018b}. In general, the applicability and success of semi-supervised methods to a specific task is not certain. It has been recently argued, and experimentally demonstrated, that the gains that have been reported by many semi-supervised methods may need to be reassessed \cite{oliver2018}.

In general, it is much cheaper and faster to obtain rough segmentations, either manual or computer-generated, on large training datasets. However, rather than treating such approximate segmentation labels as ground truth, as done in \cite{ghafoorian2018}, one can use more intelligent methods. A comprehensive recent review of deep learning with noisy labels with a focus on medical image analysis can be found in \cite{karimi2019noisylabel}. Several recent studies have reported successful applications of such methods for medical image segmentation \cite{karimi2020learning,mirikharaji2019,fries2019}.

\subsection{Model calibration and uncertainty estimation}

All machine learning models are bound to make wrong predictions on a fraction of test data. Nonetheless, one would like the confidence of predictions to be proportional to the probability of being correct. Consider a test set of $\{x_i, y_i \}_{i=1:N}$ and suppose for sample $x_i$ the model predicts the class $\hat{y}_i$ with a probability $\hat{p}_i$. In the ideal scenario with perfect confidence calibration, $\text{P}(\hat{y}=y  | \hat{p}=p)= p$ \cite{zadrozny2001}. 

Standard deep learning models have been shown to be poorly calibrated \cite{guo2017}. This should be concerning for safety-critical applications including medicine. A range of methods have been proposed for improving the calibration of deep learning models. For example, it has been shown that calibration can be improved by using a proper scoring rule as the loss function \cite{lakshminarayanan2017,guo2017}, using weight decay and \textit{avoiding} batch normalization \cite{guo2017}. Training on adversarial examples \cite{szegedy2013} has also been shown to improve model calibration \cite{lakshminarayanan2017}. Some studies have used the Platt scaling for improving the model calibration \cite{kuleshov2018,guo2017}. In \cite{guo2018}, for instance, after the deep learning network is trained, a model $q= \text{Softmax}(a z + b)$ with parameters $a$ and $b$ is trained on the logit vector, $z$, of the trained network to obtain a more calibrated prediction $q$. Another study proposed to train a separate model, to map the uncalibrated output of a CNN to calibrated probabilities \cite{maronas2019}. For this purpose, they used a Bayesian neural network, which they trained after training the main deep learning model.

Prediction uncertainty has also received some attention in medical image segmentation studies. Some studies have proposed methods to estimate the uncertainty \cite{wang2019d} or to use the prediction uncertainty to improve the segmentation accuracy \cite{karimi2019accurate,karimi2018accurate}. However, little attention has been paid to methods for \textit{improving} the calibration of CNN-based segmentation models. An example of the latter is the work of \cite{mehrtash2019}, where the authors use model ensembles to arrive at better-calibrated models. That study trained an ensemble of CCNs with random initialization of network weights and random shuffling of training data. They show that the ensemble average is better calibrated than prediction of a single method. However, their proposed method requires training and maintaining as many as 50 separate models, which is quite inefficient for many clinical applications.

\subsection{Detecting out-of-distribution data and model failure}

Another important problem in deep learning is detection of OOD data at test time. Suppose that the training data come from a distribution $x_{\text{train}} \sim D_{\text{train}}$. A central assumption of every machine learning method is that the test data come from the same distribution. When a data sample comes from an entirely different distribution than $D_{\text{train}}$, there is no performance guarantee. Ideally, the model should include a mechanism to detect the OOD data samples and issue a warning. However, this has proven to be challenging with deep learning models because of the black-box nature of these models and the highly complex mapping between their input and output. 

It has been shown that advancements in network architecture design have not improved the robustness of deep learning models to OOD data \cite{hendrycks2019}. Some studies have proposed methods that increase the robustness of deep learning models to OOD data. As an example, one study showed that some simple techniques such as histogram equalization and Adversarial Logit Pairing \cite{kannan2018} may improve robustness to perturbed and corrupted data. However, they noted that methods that work well on specific datasets may fail on other datasets. More importantly, these methods usually focus on in-distribution data that have been slightly perturbed, on which the model performance can be sub-optimal, and do not address the OOD samples, on which the model fails completely. Robustness against true OOD data has no meaning, and such data should be detected and reported/rejected.

Several studies have proposed methods for detecting OOD data in deep learning. For image classification, one study proposed training Gaussian discriminant models on the penultimate layer of the network and using the Mahalanobis Distance to detect OOD data \cite{lee2018b}. Another work suggested using the distribution of features in different layers of a deep learning model for OOD detection \cite{papernot2018}. The intuition behind that method is that if a test example is in-distribution, the training examples that are most similar to it, in terms of feature similarity, are consistent across layers. Such methods may be effective for natural image classification, where the sizes of the feature vectors is only a few hundreds or around a thousand at most and the number of training images can be millions. However, they cannot be used for 3D medical image segmentation, where feature maps are much larger and typically only tens of training images are available. A number of studies have proposed to detect OOD data based on measures of prediction uncertainty, which is usually quantified as a function of entropy of the predicted class probability \cite{kendall2017,hendrycks2016}. However, such methods have been shown to have a low accuracy in semantic segmentation applications \cite{bevandic2018}.

Compared with image classification, OOD detection in semantic image segmentation has received much less attention. A recent study found that methods proposed for OOD detection in image classification do not translate well to image segmentation tasks \cite{angus2019}. For semantic segmentation of street view images, one study proposed a dedicated neural network to detect OOD data \cite{bevandic2018}. Their approach aims to classify an image as in-distribution or OOD using a very large ``background dataset" to represent the distribution of the variety of visual scenes outside of the training data distribution. The authors use the ILSVRC dataset as the background dataset. However, it is difficult to obtain or even define the background set, especially in medical imaging. One study used prediction uncertainty measures to identify OOD data in medical image segmentation \cite{mehrtash2019}. However, they evaluated this method on data that were hard to segment, not on true OOD data. As we show in Section \ref{results_and_discussion} of this paper, methods based on prediction uncertainty cannot accurately detect OOD data.

We should mention in passing that a related topic to OOD detection is the topic of adversarial examples \cite{szegedy2013}. These are examples that are intentionally crafted to fool a model into making wrong predictions. Adversarial examples may be important in some medical applications, but they are beyond the scope of this paper, which focuses on natural OOD data.

\subsection{Contributions of this work}

In this paper, we address the critical problems discussed above and make the following contributions. 

\begin{itemize}

\item We show that multi-task learning is an effective approach for training CNNs for medical image segmentation. Specifically, rather than training a CNN to segment a single organ in a single imaging modality (e.g., prostate in MRI), we propose training a model that segments several different organs in different imaging modalities. We show that successful multi-task learning does not need any changes to the network architecture or training procedures. The network can learn to automatically recognize the context (i.e., the imaging modality and organ) and accurately segment the organ of interest without any extra input or supervision.

\item We show that multi-task learning can lead to segmentation accuracy on par with or even better than competing methods such as transfer learning. We further show, for the first time, that multi-task learning also improves the model's confidence calibration.

\item For detecting OOD test data, we devise a novel method based on spectral analysis of the CNN feature maps. We show that this method can detect OOD test data much more accurately than recently proposed methods that are based on prediction uncertainty.

\end{itemize}

\section{Materials and Methods}

\subsection{Data}

A large number of datasets were used in this study. We provide a summary of the information about these datasets in Table \ref{table:data}. Unless otherwise stated, we used 70\% of each dataset for training and validation and 30\% for test. All Computed Tomography (CT) images were normalized by a simple linear mapping that mapped the Hounsfield Unit values in the range $[-1000,1000]$ to intensity range $[0,1]$. All Magnetic Resonance (MR) images were normalized by dividing the image by the standard deviation of the voxel intensities.

\begin{table*}[!htb]
\footnotesize

 \caption{\small{Summary of the information on the datasets used in this study. The first column shows the names that we use to refer to each dataset throughout this paper.}}
  \label{table:data}
  
  \begin{center}
    \begin{tabular}{ L{2.8cm}  C{2.0cm} C{3.0cm} C{1.0cm} C{6.5cm}}
\hline
name & modality & organ & data size & source   \\ \hline
CP- younger fetus & T2 MRI & brain cortical plate & 27 &  In-house (Boston Children's Hospital)\\
CP- older fetus & T2 MRI & brain cortical plate &  15 &   In-house (Boston Children's Hospital)\\
CP- newborn & T2 MRI & brain cortical plate & 400 & \cite{bastiani2019} \\
Liver-CT & CT & liver & 19 & \cite{heimann2009b} \\
Liver-MRI-SPIR  & MRI & liver & 20 & \cite{kavur2019} \\
Liver-MRI-DUAL-in  & MRI & liver & 20 & \cite{kavur2019} \\
Liver-MRI-DUAL-out  & MRI & liver & 20 & \cite{kavur2019} \\
Heart & MRI & left atrium  & 20 & \url{https://decathlon-10.grand-challenge.org/} \\
Prostate & MRI & prostate & 32 & \url{https://decathlon-10.grand-challenge.org/} \\
Pancreas & CT & pancreas & 281 & \url{https://decathlon-10.grand-challenge.org/} \\
Hippocampus & MRI & hippocampus & 260 & \url{https://decathlon-10.grand-challenge.org/} \\
Spleen & CT & spleen & 41 & \url{https://decathlon-10.grand-challenge.org/} \\
\hline
\end{tabular}
  \end{center}

\end{table*}

\subsection{Network architecture and training details}
\label{training_details}

We used a network similar to the 3D U-Net \cite{cciccek2016}, which we modified by adding residual connections with short and long skip connections. We set the number of features in the first stage of the encoder part of the network to 14, which was the largest possible on our GPU memory. The model worked on $96\times96\times96$-voxel image blocks. During training, we sampled blocks from random locations in the training images. Other data augmentation methods used during training included flip, rotation by integer multiples of $\pi/2$, and addition of random Gaussian noise to voxel intensity values. We also experimented with elastic deformation for data augmentation (\cite{milletari2016,karimi2018segmentation}), but we did not pursue that augmentation method because it negatively impacted segmentation of fine structures such as those in the brain cortical plate. On a test image, a sliding window approach with a 24-voxel overlap between adjacent blocks was used to process the image. We used the negative of the Dice Similarity Coefficient (DSC) between the predicted and target probability maps as the loss function and Adam \cite{kingma2014} as the optimization method. We used an initial learning rate of $10^{-4}$, which was reduced by 0.90 after every 2000 training iterations if the loss did not decrease. If the loss did not decrease for two consecutive evaluations, we stopped the training and claimed convergence. This typically occurred after 100-150 training epochs through all training images.

Since the focus of the study is on the training data, model calibration, and OOD detection, we used the same settings mentioned above in all experiments. Admittedly, this may reduce the model accuracy by a small percentage because one can always choose better model size, architecture, or learning rate using cross-validation to achieve slightly better results for a specific dataset. Nonetheless, using the same setting allowed us to remove the effect of these confounding factors and focus on the factors that were the focus of our study.

\subsection{Multi-task learning}

The common practice in training CNNs for medical image segmentation is to train a CNN to segment a single organ in a single imaging modality. As we explained above, to cope with the small size of medical image datasets, methods such as transfer learning have also become common. Here, on the other hand, we advocate for multi-task learning, i.e., training on heterogeneous data. Simply, we train a single model on a mix of training datasets that can come from different imaging modalities with different organs of interest to be segmented, such as the datasets shown in Table \ref{table:data}.

We do not use additional inputs to inform the model of the image modality or the organ that needs to be segmented. Furthermore, we use the same loss function and optimization procedure. In other words, nothing changes compared with training on a single dataset. The only point worth mentioning is the frequency of sampling from different training datasets when their sizes are very different. We sample from each dataset with a probability proportional to the inverse of the square root of dataset size, $1/\sqrt{n}$. This way, if for example we train on two datasets with 10 and 100 images each, the probability of sampling an image from these two datasets will be 0.24 and 0.76, respectively. In our experience, this strategy strikes a good balance in terms of the test performance of the model on different datasets when the training dataset sizes are very different. Using $1/n$ instead of $1/\sqrt{n}$, for example, resulted in poor performance on datasets with smaller number of images.

\subsection{OOD detection based on the spectral signature of feature maps}
\label{OOD_detection_method}

We propose a novel method for detecting OOD data samples that are input to a CNN-based medical image segmentation model. As we mentioned above, such models produce over-confident predictions even when a test sample is entirely different from the training data. For example, a network trained on the Liver-CT dataset produces confident (but, obviously, completely wrong) segmentations on the brain cortical plate. As we show in Section \ref{results_and_discussion}, even on such seemingly simple cases, previously-proposed methods based on prediction uncertainty are unable to accurately detect model failure.

Due to the large size of 3D medical images and their computed features, a method based on analyzing the feature maps or the predicted segmentation map in their native space is unclear and likely to be ineffective. Instead, we propose computing the spectrum of the feature maps, which we define as the vector of singular values computed using a singular value decomposition (SVD). Consider a test image $x_i$ and denote the feature map computed for this image at a certain stage (i.e., layer) of the network with $F_i \in \mathbb{R}^{w, h, d, n}$, where $w, h, d$ denote the dimensions of the feature map and $n$ is the number of features. We reshape $F_i$ as $\mathbb{R}^{whd, n}$ and compute the SVD of $F_i$ as $F_i= U S V$, where $U$ and $V$ are orthonormal matrices and the diagonal matrix $S$ contains the singular values of $F_i$, which is referred to as its spectrum \cite{golub2013}. Values of the vector of singular values $s= \text{diag}(S)$ depend on the magnitude of the feature values, which in turn depend on the image voxel intensities. Moreover, the spectrum has a very large dynamic range. To eliminate these effects, we take the logarithm of the spectrum $s$ and then normalize it so that it has an $\ell_2$ norm of unity. We refer to the normalized logarithmic spectrum of the feature maps computed as explained above as ``the spectral signature" of the feature maps corresponding to an organ of interest. We still denote this spectral signature with $s$ in the following.

\begin{figure}[!htb]
  \centering
  \centerline{\includegraphics[width=90mm]{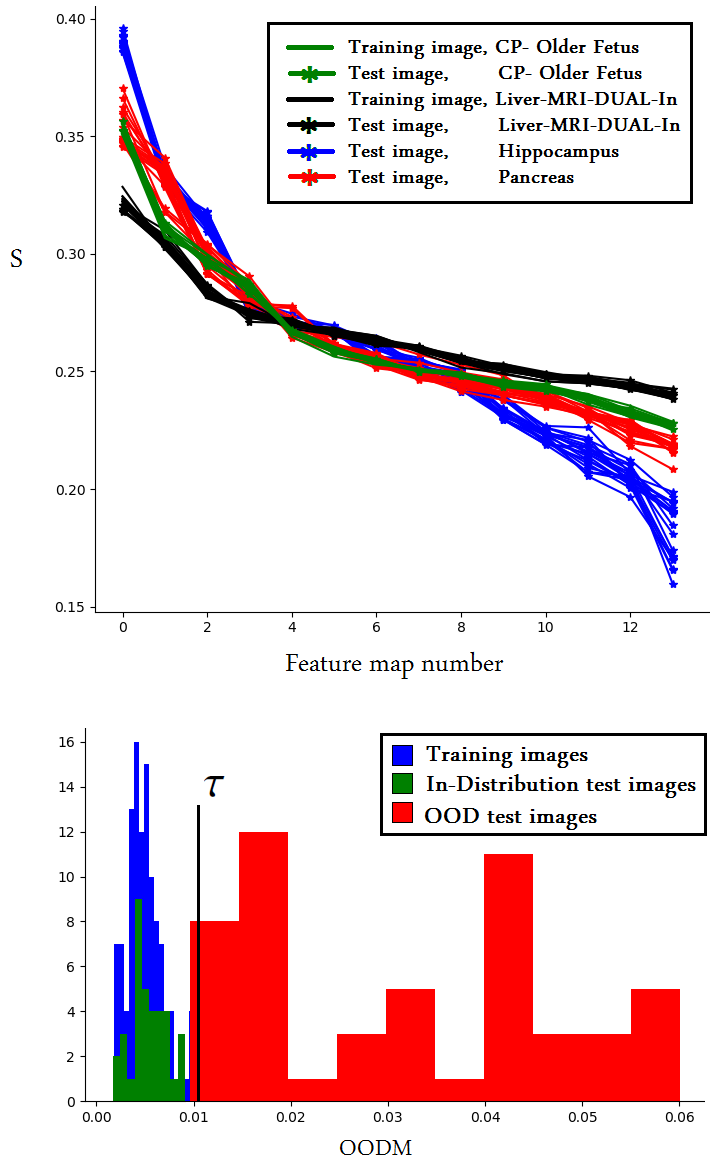}}
\caption{A demonstration of our proposed OOD detection method in action. These figures were generated from a model that was trained on eight datasets including (See Table \ref{table:data}): CP- younger fetus, CP- older fetus, Prostate, Heart, Liver-CT, Liver-MRI-SPIR, Liver-MRI-DUAL-In and Liver-MRI-DUAL-Out. TOP: Spectral signatures of feature maps for four different datasets. Two of these datasets (i.e., CP- older fetus and Liver-MRI-DUAL-In) are from the distribution of the training images, while the other two (i.e., Hippocampus and Pancreas) are OOD. We have shown the spectra for only four datasets in order to avoid clutter. BOTTOM: Histograms of OODM values (computed using Eq. \eqref{eq:OODM}) for training images, in-distribution test images (i.e., test images from the same eight datasets as the training images), and OOD test images. OOD test images are from Pancreas, Hippocampus, and Spleen datasets. The value of the threshold $\tau= 0.011$ has been marked with the vertical black line.}
\label{fig:signature}
\end{figure}

In Figure \ref{fig:signature}, we have shown examples of how these signatures look like. This figure is for a model trained on several datasets from Table \ref{table:data} including CP- younger fetus and Liver-MRI-SPIR datasets but not including Pancreas and Hippocampus datasets. The figure shows example spectral signatures of training images from these four datasets. Clearly, each dataset has a distinct spectral signature. Note that this model segments CP- younger fetus and Liver-MRI-SPIR accurately, but fails completely on Pancreas and Hippocampus, which have not been seen during training. Nonetheless, as we show in Section \ref{results_and_discussion}, methods based on uncertainty measures cannot detect these as OOD.

We suggest detecting OOD data based on the dissimilarity of the spectral signatures. For all images in the training data, $X_{\text{train}}$, we compute their spectral signatures and save them in a matrix, $S_{\text{train}}$. Given a test image, $x_{\text{test}}^i$, we compute its spectral signature $s_{\text{test}}^i$. We then compare $s_{\text{test}}^i$ with the spectral signature of the training data by computing Out-Of-Distribution Measure (OODM), which we define as:

\begin{equation} \label{eq:OODM}
\text{OODM}(x_{\text{test}}^i)= \min_j \left( \| s_{\text{test}}^i - s_{\text{train}}^j  \|_2 \hspace{2mm} , \hspace{2mm}  s_{\text{train}}^j \in S_{\text{train}} \right)
\end{equation}

\noindent In other words, $\text{OODM}(x_{\text{test}}^i)$ is the Euclidean distance of the spectral signature of $x_{\text{test}}^i$ to its nearest neighbor in the training set.

We anticipate that for test images coming from the distribution of the training data, $\text{OODM}$ should be smaller than for images coming from other distributions. We declare a test image $x_{\text{test}}^i$ to be OOD if $\text{OODM}(x_{\text{test}}^i)>\tau$. The threshold $\tau$ is determined using the training data. Specifically, on the training data we compute the vector of $\text{OODM}_{\text{train}}$ using Eq. \eqref{eq:OODM} on a leave-one-out basis, i.e., by comparing the spectrum of each training image with the spectra of all other training images. We then compute $\tau$ as:

\begin{equation} \label{eq:tau}
\tau= \text{mean}(\text{OODM}_{\text{train}})+C \times  \text{std}(\text{OODM}_{\text{train}}),
\end{equation}

\noindent where we set $C=2.5$ for computing the detection accuracy.

Deep learning models compute a large number of feature maps from an input image. In practice, one can compute the spectral signature on any/all feature map(s). However, we found that using deepest feature maps leads to better results for the purpose of OOD detection. This is in agreement with the known fact that deeper layers provide more disentangled manifolds \cite{bengio2013b,feinman2017}. In this study, we only worked with the very last (i.e., deepest) feature maps. In our network, the number of channels in this feature map was 14, which was the length of the spectral signatures in this work.

Figure \ref{fig:signature} shows an example of histograms of $\text{OODM}$ values for the training data, in-distribution test data, and OOD test data. The histograms show that the proposed OODM easily separates in-distribution from OOD data in this experiment. 

In this study, we compared our proposed OOD detection method with a common strategy based on prediction uncertainty \cite{kendall2017}. Specifically, we trained our models using dropout (with a rate of 10\%) after all convolutional layers. At test time, we drew $N=10$ random dropout masks and computed the average of these segmentation probability maps. We used the entropy of this mean probability map, $H(\bar{p})= - \bar{p} \log (\bar{p})$ as an estimated voxel-wise map of prediction uncertainty. To estimate an image-wise uncertainty, as suggested in \cite{mehrtash2019}, we used the average of the voxel-wise uncertainty on the predicted foreground. Similar to our approach with $\text{OODM}$ explained above, we computed a threshold similar to Eq. \eqref{eq:tau} on the training set. This threshold was used to determine if a test image was OOD.

\subsection{Evaluation metrics}

We quantify segmentation accuracy using DSC, the 95 percentile of the Hausdorff Distance (HD95) and Average Symmetric Surface Distance (ASSD). To assess model calibration, we compute the Expected Calibration Error (ECE) and Maximum Calibration Error (MCE), as proposed in \cite{naeini2015}. For OOD detection experiments, we report accuracy, sensitivity, specificity. We also compute the area under the receiver-operating characteristic curve (AUC) by changing the value of $\tau$.

\section{Results and Discussion}
\label{results_and_discussion}

\subsection{Feasibility and benefits of multi-task learning}

As we mentioned above, we propose training a single model to segment different organs in different imaging modalities. To show that this is a viable approach, we trained a model on seven datasets spanning six different organs in MRI and CT images. We then trained seven separate models, one on each of these seven datasets. We show a comparison of the test performance of these two training strategies in Table 2.

\begin{table*}[!htb]
\footnotesize

    \label{table:heterogeneous_training_table}
 \caption{\small{Results of an experiment to compare training a single model on several datasets with training dedicated models separately for each dataset. This experiment included seven different datasets representing six different organs in MRI and CT.}}
 
\begin{tabular}{ L{4.0cm}  L{3.0cm} C{1.6cm} C{1.6cm} C{1.6cm} C{1.6cm} C{1.6cm} }
\thickhline
 
Training method & Data & DSC & HD95 (mm) & ASSD (mm) & ECE & MCE   \\ \thickhline

\multirow{7}{*}{\parbox{3cm}{Training a separate model for each dataset}} & CP- younger fetus  & $\bm{0.90 \pm 0.03}$ & $0.80 \pm 0.02$ & $\bm{0.22 \pm 0.03}$ & $0.08 \pm 0.02$ & $0.21 \pm 0.05$  \\
& CP- older fetus  & $0.82 \pm 0.05$ & $1.02 \pm 0.19$ & $0.36 \pm 0.09$ & $0.13 \pm 0.05$ & $0.31 \pm 0.12$  \\
& Heart  & $\bm{0.90 \pm 0.04}$ & $\bm{9.83 \pm 15.5}$ & $\bm{1.74 \pm 1.86}$ & $0.19 \pm 0.04$ & $0.35 \pm 0.10$  \\
& Hippocampus  & $\bm{0.88 \pm 0.02}$ & $\bm{1.01 \pm 0.23}$ & $\bm{0.44 \pm 0.07}$ & $\bm{0.17 \pm 0.02}$ & $\bm{0.34 \pm 0.06}$  \\
& Prostate  & $0.85 \pm 0.05$ & $14.8 \pm 24.5$ & $3.7 \pm 4.7$ & $0.27 \pm 0.07$ & $0.39 \pm 0.11$  \\
& Liver-CT  & $\bm{0.97 \pm 0.01}$ & $5.07 \pm 1.94$ & $1.47 \pm 0.33$ & $0.11 \pm 0.01$ & $0.26 \pm 0.02$  \\
& Liver-MRI-SPIR  & $0.90 \pm 0.01$ & $32.0 \pm 20.3$ & $5.86 \pm 1.68$ & $\bm{0.20 \pm 0.03}$ & $\bm{0.36 \pm 0.06}$  \\

\thickhline

\multirow{7}{*}{\parbox{3cm}{Training a single model for all datasets}} & CP- younger fetus  & $0.89 \pm 0.04$ & $\bm{0.80 \pm 0.01}$ & $0.23 \pm 0.04$ & $\bm{0.07 \pm 0.01}$ & $\bm{0.20 \pm 0.04}$  \\
& CP- older fetus  & $\bm{0.84 \pm 0.02}$ & $\bm{0.91 \pm 0.19}$ & $\bm{0.34 \pm 0.06}$ & $\bm{0.12 \pm 0.02}$ & $\bm{0.26 \pm 0.05}$  \\
& Heart  & $0.88 \pm 0.08$ & $11.0 \pm 19.5$ & $3.48 \pm 5.26$ & $\bm{0.13 \pm 0.02}$ & $\bm{0.24 \pm 0.08}$  \\
& Hippocampus  & $0.87 \pm 0.02$ & $1.26 \pm 0.22$ & $0.50 \pm 0.07$ & $0.18 \pm 0.02$ & $0.35 \pm 0.07$  \\
& Prostate  & $\bm{0.88 \pm 0.06}$ & $\bm{5.61 \pm 2.78}$ & $\bm{1.96 \pm 0.68}$ & $\bm{0.25 \pm 0.06}$ & $\bm{0.42 \pm 0.11}$  \\
& Liver-CT   & $\bm{0.97 \pm 0.01}$ & $\bm{4.58 \pm 1.01}$ & $\bm{1.40 \pm 0.36}$ & $\bm{0.10 \pm 0.02}$ & $\bm{0.20 \pm 0.04}$  \\
& Liver-MRI-SPIR  & $\bm{0.92 \pm 0.02}$ & $\bm{13.0 \pm 7.91}$ & $\bm{3.77 \pm 1.68}$ & $0.21 \pm 0.05$ & $0.38 \pm 0.06$  \\

\thickhline

\end{tabular}

\end{table*}

The results are very interesting. They show that training a single model for several different datasets can achieve results that are as good as or even better than when dedicated models are trained separately for each dataset. In terms of segmentation accuracy, a joint model trained on heterogeneous data was overall better than models dedicated to a single dataset. When a dedicated model was better than the joint model, the difference was small, typically within 10\%. On the other hand, on some datasets the joint model improved the segmentation accuracy by large margins. For example, on Prostate and Liver-MRI-SPIR datasets, the joint model reduced HD95 and ASSD by factors of 1.55-2.64, which is quite substantial. The joint model was also better-calibrated than the dedicated models on 5 out of 7 datasets. Only on the Hippocampus dataset, the dedicated model was noticeably better than the joint model. It is interesting to note that the Hippocampus dataset included 260 images, compared with 15-32 images in each of the other six datasets used in this experiment. This indicates the influence of dataset size on the potential benefits of training on heterogeneous data.

\begin{figure*}[!htb]
\begin{minipage}[b]{1.0\linewidth}
  \centering
  \centerline{\includegraphics[width=\textwidth]{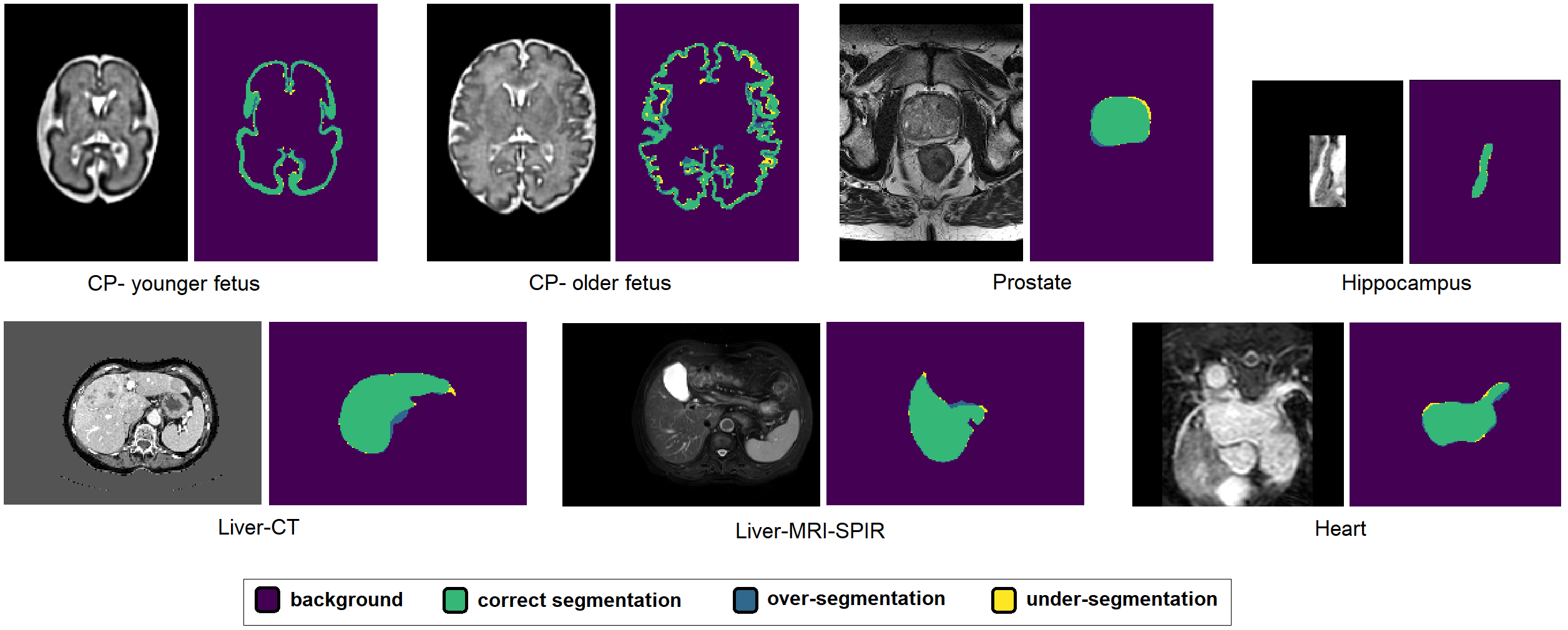}}
\end{minipage}
\caption{A slice of a test image from each of the seven datasets used in the experiment reported in Table 2 and the output segmentation of the joint model that was trained on all seven datasets. This single joint model trained on all seven datasets was able to accurately segment different organs in different modalities. Moreover, it performed as well as or even better than seven dedicated models trained to segment each dataset separately.}
\label{fig:segment_everything}
\end{figure*}

In retrospect, given the small size of most of our datasets, the fact that a single model can automatically recognize the context and accurately segment the organ of interest is interesting. Figure \ref{fig:segment_everything} shows a slice of one test image from each of the seven datasets used in this experiment and the segmentation produced by the joint model trained on all seven datasets. The model accurately segments all seven datasets. In other experiments, we increased the number of datasets to 12, and we observed the same patterns as those shown for the experiment with seven datasets in Table 2.

\begin{figure*}[!htb]
\begin{minipage}[b]{1.0\linewidth}
  \centering
  \centerline{\includegraphics[width=\textwidth]{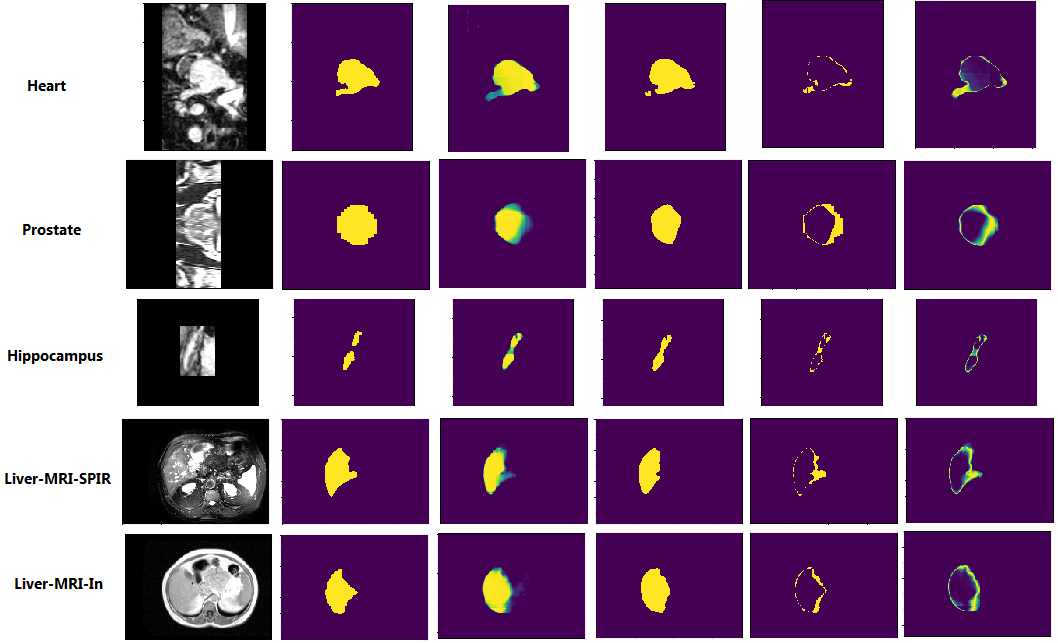}}
\end{minipage}
\caption{Examples of prediction uncertainty maps produced by a model trained to segment a heterogeneous pool of datasets. From left, the first column shows a slice of the image. The second column is the ground-truth segmentation map. The third column is the probability map (in the range [0,1]) that each voxel is a foreground voxel. The fourth column is the probability map thresholded at 0.5, showing the binary prediction of the model. The fifth column shows the binary difference between the ground-truth (second column) and prediction (fourth column). The last column shows the estimated prediction uncertainty map (in the range [0, $-0.5 \log(0.5)$]).}
\label{fig:uncertainty}
\end{figure*}

Our quantitative evaluation of model calibration in this experiment, and throughout the paper, is based on ECE and MCE. Table 2 shows that in general multi-task learning improves the model prediction calibration.  Again, one of the exceptions was the Hippocampus dataset, for which 260 training images were available. Figure \ref{fig:uncertainty} shows examples of estimated uncertainty maps. For this figure, we have intentionally chosen example test images on which the segmentation accuracy was relatively low. The figure shows that the model displays high segmentation uncertainty at the locations where segmentation error occurs, visually confirming that the model is well-calibrated.

\begin{figure}[!htb]
\begin{minipage}[b]{1.0\linewidth}
  \centering
  \centerline{\includegraphics[width=90mm]{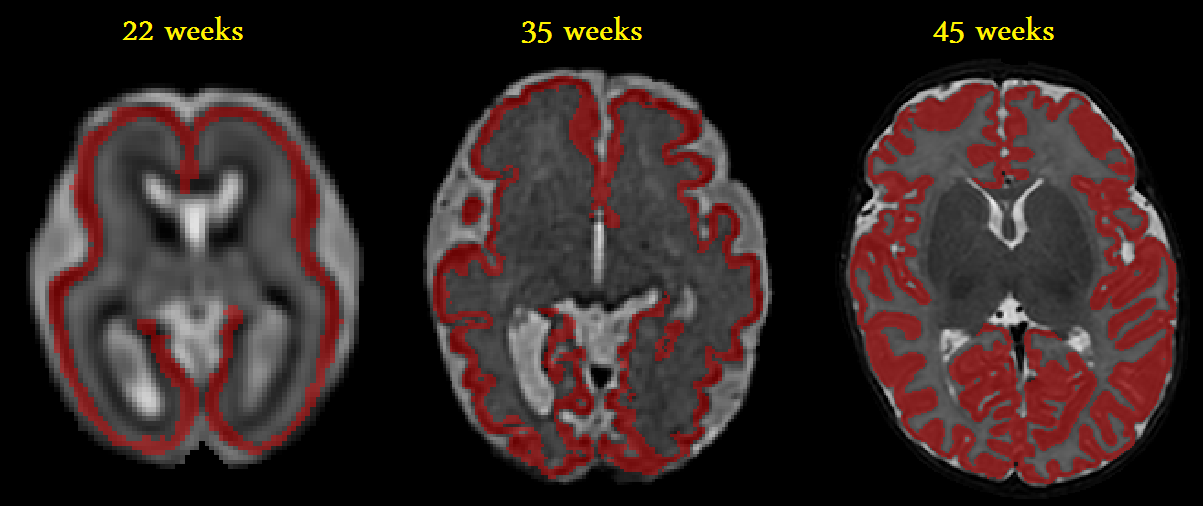}}
\end{minipage}
\caption{Example axial slices of the images and segmentations from the three cortical plate segmentation datasets used in this study. From left to right, the images come from CP- younger fetus, CP- older fetus, and CP- newborn. Postmenstrual age of each subject is displayed above the image.}
\label{fig:cp_samples}
\end{figure}

In order to show an important potential benefit of the above-described multi-task learning approach, we compare it with transfer learning in an experiment involving the three cortical plate datasets (See Table \ref{table:data}). As shown in the example images and segmentations in Figure \ref{fig:cp_samples}, the shape and complexity of cortical plate evolves dramatically before and after birth. In addition, the sizes of the three datasets are highly unequal. CP- younger fetus dataset includes 27 images with post-menstrual age of $25.1 \pm 3.84$ weeks, CP- older fetus includes 15 images with age of $33.7 \pm 1.80$ weeks, and CP- newborn includes 400 images with age of $39.8 \pm 2.94$ weeks. The question is, given the complexity of this segmentation task and the small size of two of the datasets, what is the best training strategy to achieve high segmentation accuracy on all three datasets?

Given the much smaller sizes of two of the datasets, transfer learning is the method that is recommended by previous studies (\cite{tajbakhsh2016, cheplygina2019}). In Table 3, we compare the results obtained using different transfer learning trials with the results obtained using multi-task learning on heterogeneous data, i.e., training a single model on all three datasets. In each of the transfer learning trials, we first trained the model to convergence on one of the datasets. We then fine-tuned it to convergence on another dataset. We then further fine-tuned the model that had been trained on the second dataset on the remaining dataset. Our definition of convergence is the same as defined in Section \ref{training_details}. Our fine-tuning strategy was ``deep fine-tuning" \cite{tajbakhsh2016}; we reduced the initial learning rate by half and fine-tuned all model layers. We also experimented with shallow fine-tuning as well as keeping the initial learning rate, but the results were inferior.

\begin{table*}[!htb]
\footnotesize

    \label{table:cp_finetune}
 \caption{\small{Results of experiments on cortical plate segmentation. We compare three different transfer learning approaches with our proposed method of multi-task learning with heterogeneous data, i.e., training a single model to segment all three datasets. For each of the three datasets, we have highlighted the best results using bold type.}}
 
\begin{tabular}{ L{5.0cm}  L{2.4cm} C{1.5cm} C{1.5cm} C{1.5cm} C{1.5cm} C{1.5cm} }
\thickhline
 
Training/fine-tuning data & Test data & DSC & HD95 & ASSD & ECE & MCE   \\ \thickhline

Train on CP- younger fetus & CP- younger fetus  &  $\bm{0.90 \pm 0.03}$ & $0.80 \pm 0.02$ & $0.22 \pm 0.03$ & $0.08 \pm 0.02$ & $0.21 \pm 0.05$  \\[5pt]
\hline
 \hspace{4mm} \mbox{\Large $\rotatebox[origin=c]{180}{$\Lsh$}$} Fine-tune on CP- older fetus & CP- older fetus  & $0.80 \pm 0.06$ & $1.02 \pm 0.19$ & $0.38 \pm 0.11$ & $0.17 \pm 0.06$ & $0.37 \pm 0.10$  \\[5pt]
\hline
\hspace{8mm} \mbox{\Large $\rotatebox[origin=c]{180}{$\Lsh$}$} Fine-tune on CP- newborn & CP- newborn  & $0.93 \pm 0.07$ & $0.80 \pm 0.02$ & $\bm{0.16 \pm 0.01}$ & $0.09 \pm 0.01$ & $0.23 \pm 0.02$  \\[5pt]
\thickhline

Train on CP- older fetus & CP- older fetus  & $0.82 \pm 0.05$ & $1.02 \pm 0.19$ & $0.36 \pm 0.09$ & $0.13 \pm 0.05$ & $0.31 \pm 0.12$  \\[5pt]
\hline
 \hspace{4mm} \mbox{\Large $\rotatebox[origin=c]{180}{$\Lsh$}$} Fine-tune on CP- younger fetus & CP- younger fetus  & $\bm{0.90 \pm 0.03}$ & $0.83 \pm 0.03$ & $\bm{0.20 \pm 0.03}$ & $0.10 \pm 0.03$ & $0.22 \pm 0.07$  \\[5pt]
\hline
\hspace{8mm} \mbox{\Large $\rotatebox[origin=c]{180}{$\Lsh$}$} Fine-tune on CP- newborn & CP- newborn  & $\bm{0.93 \pm 0.01}$ & $0.85 \pm 0.02$ & $\bm{0.16 \pm 0.01}$ & $0.08 \pm 0.01$ & $0.23 \pm 0.03$ \\[5pt]
\thickhline

Train on CP- newborn & CP- newborn  
& $0.92 \pm 0.01$ & $0.82 \pm 0.02$ & $0.19 \pm 0.01$ & $0.05 \pm 0.01$ & $0.16 \pm 0.03$  \\[5pt]
\hline
 \hspace{4mm} \mbox{\Large $\rotatebox[origin=c]{180}{$\Lsh$}$} Fine-tune on CP- younger fetus & CP- younger fetus  & $0.90 \pm 0.03$ & $\bm{0.80 \pm 0.01}$ & $\bm{0.20 \pm 0.03}$ & $0.10 \pm 0.02$ & $0.23 \pm 0.05$
\\[5pt]
\hline
\hspace{8mm} \mbox{\Large $\rotatebox[origin=c]{180}{$\Lsh$}$} Fine-tune on CP- older fetus & CP- older fetus  & $0.82 \pm 0.05$ & $\bm{0.91 \pm 0.19}$ & $0.35 \pm 0.10$ & $0.19 \pm 0.05$ & $0.38 \pm 0.08$   \\[5pt]
\thickhline

\multirow{4}{*}{\parbox{5cm}{Train a single model for all datasets}} & CP- younger fetus  &  $\bm{0.90 \pm 0.03}$ & $\bm{0.80 \pm 0.01}$ & $0.22 \pm 0.02$ & $\bm{0.06 \pm 0.03}$ & $\bm{0.18 \pm 0.09}$  \\[5pt]
& CP- older fetus & $\bm{0.85 \pm 0.03}$ & $\bm{0.91 \pm 0.19}$ & $\bm{0.32 \pm 0.07}$ & $\bm{0.04 \pm 0.02}$ & $\bm{0.09 \pm 0.06}$  \\[5pt]
& CP- newborn & $0.92 \pm 0.01$ & $\bm{0.80 \pm 0.01}$ & $0.20 \pm 0.01$ & $\bm{0.03 \pm 0.01}$ & $\bm{0.10 \pm 0.03}$  \\ 
\thickhline

\end{tabular}

\end{table*}

The results are interesting. Transfer learning improved the segmentation accuracy in some cases, but in most cases the improvement was very small. Training a joint model on all three datasets, on the other hand, achieved segmentation accuracy results that were on par with or better than any of the transfer learning trials. For the smallest dataset, i.e., CP- older fetus, the joint model achieved the best results in terms of DSC, HD, and ASSD. Furthermore, the joint model had better-calibrated predictions than all of the three transfer learning approaches on all three datasets. 

An additional appeal of a joint model that accurately segments all three datasets is its universality. This implies that we will need to maintain only one set of model weights. On the other hand, a model that has been trained on any single one of these datasets will have a poor performance on the other datasets. Therefore, we will need to maintain three separate trained models, one for each dataset. Moreover, for a test image, we will need to know which of the three datasets the image belongs to, in order to use the right model on that image.

\subsection{Detecting OOD test data}

In this section, we present the results of our proposed OOD detection method in three different experiments and compare it with the method based on prediction uncertainty.

In the first experiment, we used a mixture of eight different datasets for training. These included CP- younger fetus, CP- older fetus, Prostate, Heart, Liver-CT, Liver-MRI-SPIR, Liver-MRI-DUAL-In and Liver-MRI-DUAL-Out datasets. Then, we applied the proposed OOD detection method on the trained model. We used test images from the same eight dataset as in-distribution data. As OOD data, we used Pancreas, Hippocampus, and Spleen datasets. Histogram of the proposed OODM for this experiment has been shown in Figure \ref{fig:signature}(b). Table \ref{table:OOD_results_1} shows comparison of our method with the method based on prediction uncertainty. Our method perfectly detected the OOD images, but the method based on prediction uncertainty failed.

\begin{table}[!htb]

  \caption{\footnotesize{Comparison of the proposed OOD detection method with the method based on prediction uncertainty. In this experiment, the in-distribution data came from CP- younger fetus, CP- older fetus, Prostate, Heart, Liver-CT, Liver-MRI-SPIR, Liver-MRI-DUAL-In and Liver-MRI-DUAL-Out datasets. The OOD data came from Pancreas, Hippocampus, and Spleen datasets.}}
  \label{table:OOD_results_1}
  
\footnotesize
  \begin{center}
    \begin{tabular}{L{2.2cm} C{1.0cm} C{1.2cm} C{1.2cm} C{0.8cm} }
\hline
Method &  accuracy &  sensitivity &  specificity & AUC \\  \hline
Proposed method     & $1.00$ & $0.98$ & $1.00$ & $1.00$ \\
Uncertainty-based     & $0.55$ & $0.48$ & $0.63$ & $0.62$ \\
\hline 
    \end{tabular}
  \end{center}

\end{table}

\begin{figure}[!htb]
  \centering
  \centerline{\includegraphics[width=90mm]{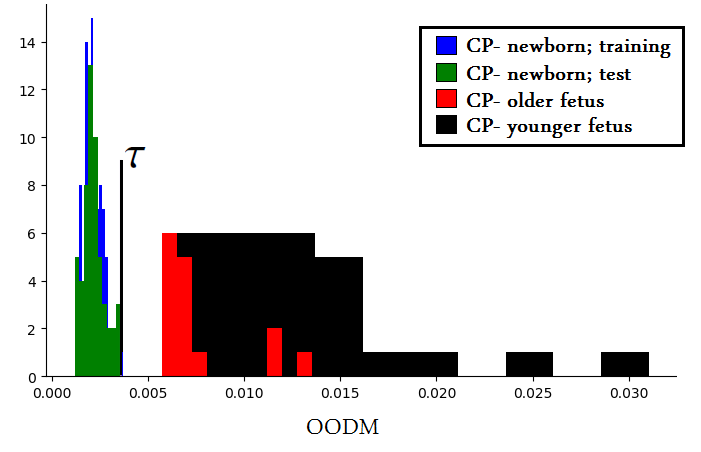}}
\caption{Histograms of OODM values (computed using Eq. \eqref{eq:OODM}) for an experiment on cortical plate segmentation. This model was trained on CP- newborn dataset.  The value of the threshold $\tau= 0.00358$ has been marked with the vertical black line.}
\label{fig:oodm_cp}
\end{figure}

In the second experiment, we trained a model on the CP- newborn dataset. We then applied it on the test data from the same dataset and on the other two cortical plate datasets. The histograms of OODM values for this experiment have been show in Figure \ref{fig:oodm_cp}. The OODM values for both CP- younger fetus and CP- older fetus fall outside of the distribution of the OODM values for CP- newborn. This model, trained only on the CP- newborn, achieved DSC values of $0.689 \pm 0.095$ and $0.781 \pm 0.028$ on the CP- younger fetus and CP- older fetus datasets, respectively. These are very low values, compared with the results shown for these datasets in Table 3. Therefore, for this model, images from both CP- younger fetus and CP- older fetus datasets should be considered as OOD. Our proposed method easily distinguished OOD data from in-distribution data. It is interesting to note that OODM values for CP- younger fetus dataset are distributed farther away, compared with those of CP- older fetus dataset. This makes sense because as shown in Figure \ref{fig:cp_samples}, CP- younger fetus is less similar to CP- newborn than CP- older fetus is. Table \ref{table:OOD_results_cp} shows comparison of our method with OOD detection based on prediction uncertainty. Compared to our method that perfectly separated in-distribution and OOD data, the method based on uncertainty prediction showed a very low accuracy.

\begin{table}[!htb]

  \caption{\footnotesize{Comparison of the proposed OOD detection method with the method based on prediction uncertainty in an experiment on cortical plate segmentation. The model is trained on CP- newborn data. The data from CP- younger fetus and CP- older fetus datasets are used as OOD data.}}
  \label{table:OOD_results_cp}
  
\footnotesize
  \begin{center}
    \begin{tabular}{L{2.2cm} C{1.0cm} C{1.2cm} C{1.2cm} C{0.8cm} }
\hline
Method &  accuracy &  sensitivity &  specificity & AUC \\  \hline
Proposed method     & $1.00$ & $1.00$ & $1.00$ & $1.00$ \\
Uncertainty-based     & $0.57$ & $0.54$ & $0.68$ & $0.67$ \\
\hline 
    \end{tabular}
  \end{center}

\end{table}

\begin{figure}[!htb]
  \centering
  \centerline{\includegraphics[width=9cm]{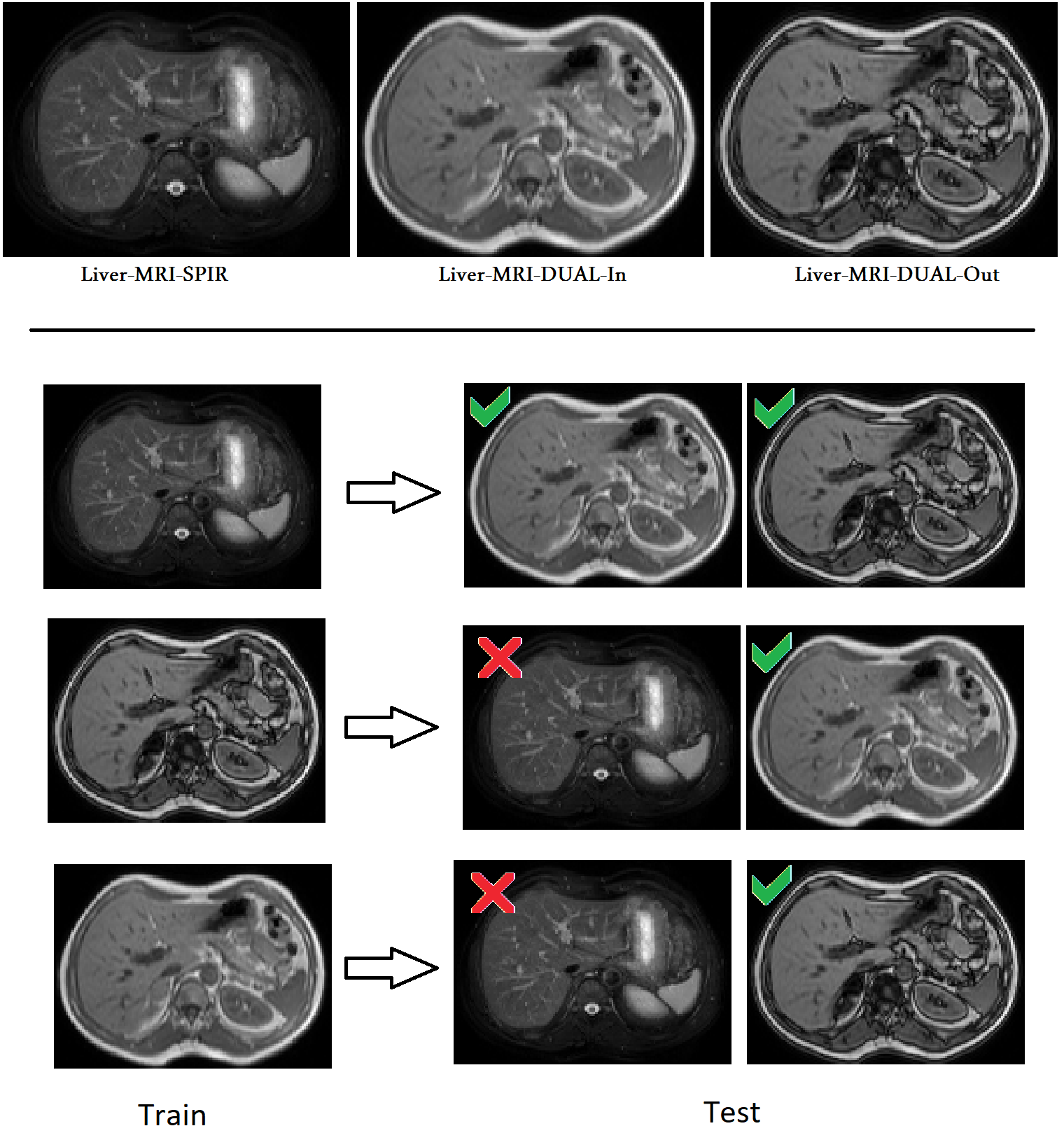}}
\caption{TOP: An axial slice of one image from each of the three liver MRI datasets. BOTTOM: A model trained on Liver-MRI-SPIR performed well on images from the other two datasets. However, a model trained on Liver-MRI-DUAL-In and/or Liver-MRI-DUAL-Out completely failed on images from Liver-MRI-SPIR. Green \cmark and red \xmark \, symbols, respectively, denote success and failure on a dataset at test time.}
\label{fig:chaos_samples}
\end{figure}

As the final experiment in OOD detection, we report the results of an experiment with the three liver MRI datasets (See Table \ref{table:data}). A slice of one sample image from each of these datasets has been shown in Figure \ref{fig:chaos_samples}. This is a very interesting example because it demonstrates that OOD data are often not easy to distinguish visually. We experimented extensively with these three datasets. We observed that when we trained a model on Liver-MRI-SPIR and Liver-MRI-DUAL-In, it segmented images from Liver-MRI-DUAL-Out with good accuracy (mean DSC= 0.89). Similarly, a model trained on Liver-MRI-SPIR and Liver-MRI-DUAL-Out, achieved a mean DSC of 0.86 on images from Liver-MRI-DUAL-In. Even a model that was trained on Liver-MRI-DUAL-SPIR alone, could segment Liver-MRI-DUAL-In and Liver-MRI-DUAL-Out images accurately. On the other hand, a model trained on Liver-MRI-DUAL-In and/or Liver-MRI-DUAL-Out failed on images from Liver-MRI-SPIR (mean DSC $\approx$ 0.40).

These observations are not intuitive, and they are not at all easy to foretell by visually inspecting these images. This example further highlights the importance of OOD detection in CNN-based medical image segmentation.

\begin{figure}[!htb]
  \centering
  \centerline{\includegraphics[width=90mm]{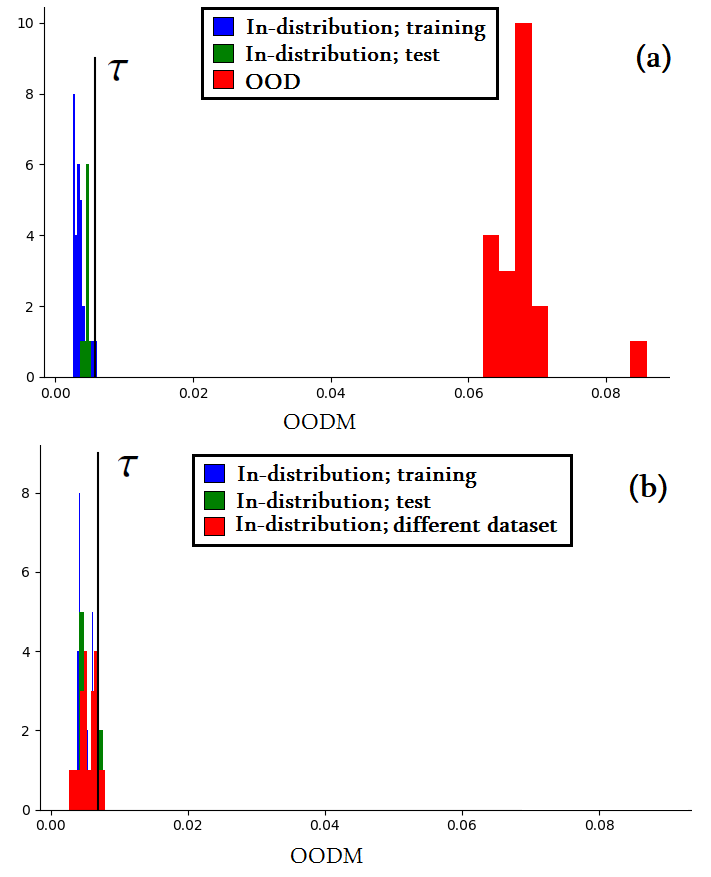}}
\caption{(a) Histograms of OODM values for an experiment on liver segmentation in MRI. The in-distribution data in this experiment included Liver-MRI-DUAL-In and Liver-MRI-DUAL-Out datasets, which were used to train the model. The OOD data included Liver-MRI-SPIR dataset, on which the model failed at test time. The value of the threshold $\tau= 0.0057$ has been marked with the vertical black line. The proposed OODM perfectly separated the OOD data from in-distribution data. (b) In this experiment, Liver-MRI-SPIR and Liver-MRI-DUAL-In were used for training. At test time, in addition to these two dataset, the model accurately segmented Liver-MRI-DUAL-Out dataset (DSC= 0.886). As can be seen, the OODM values for Liver-MRI-DUAL-Out are distributed very similar to the OODM values for the training data.}
\label{fig:oodm_liver}
\end{figure}

Figure \ref{fig:oodm_liver}(a) shows the histograms of the proposed OODM values for an experiment with these datasets. In this experiment, Liver-MRI-DUAL-In and Liver-MRI-DUAL-Out were used to train a model. The OODM values were then computed on the test data from the same two datasets as well as the data from Liver-MRI-SPIR, which are OOD for this model. The figure shows that the proposed OODM easily separates in-distribution from OOD data in this experiment. In Table \ref{table:OOD_results_liver}, we compare the proposed method with the method based on prediction uncertainty in this experiment. Similar to the two experiments presented above, the uncertainty-based method has a very low accuracy, whereas our proposed method achieves perfect detection accuracy. For completeness, Figure \ref{fig:oodm_liver}(b) shows the OODM histograms for an experiment in which Liver-MRI-SPIR and Liver-MRI-DUAL-In datasets were used for training. The trained model works well on Liver-MRI-DUAL-Out dataset as well. Therefore, all three datasets are in-distribution data. As expected, the OODM values for most images from Liver-MRI-DUAL-Out fall below the threshold $\tau$, and hence correctly classified as in-distribution.

\begin{table}[!htb]

  \caption{\footnotesize{Comparison of the proposed OOD detection method with the method based on prediction uncertainty in an experiment on liver MRI datasets. In this experiment the model was trained on Liver-MRI-DUAL-In and Liver-MRI-DUAL-Out datasets. The data from Liver-MRI-SPIR dataset are used as OOD.}}
  \label{table:OOD_results_liver}
  
\footnotesize
  \begin{center}
    \begin{tabular}{L{2.2cm} C{1.0cm} C{1.2cm} C{1.2cm} C{0.8cm} }
\hline
Method &  accuracy &  sensitivity &  specificity & AUC \\  \hline
Proposed method     & $1.00$ & $1.00$ & $1.00$ & $1.00$ \\
Uncertainty-based   & $0.64$ & $0.61$ & $0.60$ & $0.65$ \\
\hline 
    \end{tabular}
  \end{center}

\end{table}

\section{Acknowledgements}

Research reported in this publication was supported in part by the National Institutes of Health (NIH) grants R01 EB018988, R01 NS106030, and R01 EB031849; and in part by the Office of the Director of the NIH under award number S10OD0250111. The content is solely the responsibility of the authors and does not necessarily represent the official views of the NIH.

The dHCP dataset is provided by the developing Human Connectome Project, KCL-Imperial-Oxford Consortium funded by the European Research Council under the European Union Seventh Framework Programme (FP/2007-2013) / ERC Grant Agreement no. [319456]. We are grateful to the families who generously supported this trial.

\section{Conclusion}

The methods proposed in this study represent significant progress towards improving the confidence calibration and OOD detection for CNN-based medical image segmentation models. These are important contributions because they improve the reliability of these models and facilitate their wider adoption in medical and clinical settings.

We showed that standard CNN-based segmentation models can automatically recognize the context and segment the organ of interest in a large pool of heterogeneous datasets. We showed experimentally that such a joint model achieved segmentation accuracy on par with or better than dedicated models trained separately on each dataset. Our experiments also showed that models trained on heterogeneous data usually have better-calibrated predictions. These are very encouraging results. For example, as we showed in our experiment on cortical plate segmentation, this means one could train a single model to cover images from a wide rage of age groups. Not only such a model can have more accurate and better-calibrated predictions, one would need to maintain a single model that would work on all age groups, without the need to know the age of the subject at test time. Such situations are quite common in medical applications, where the available training data may show large variability in terms of subject age, body size, imaging modality, image quality, etc. While an investigation of all these factors is beyond the scope of a single study, our results show that CNN-based medical image segmentation models have the potential to handle such sources of data heterogeneity easily and effectively.

Our proposed OOD detection method can also be very valuable in practice. Whereas most previous studies have used measures of prediction uncertainty for this purpose, our experiments show that such methods can be inaccurate. To the best of our knowledge, this is the first study to propose a method for OOD detection in medical image segmentation by analyzing CNN features. In three different experiments, our proposed method based on spectral analysis of CNN feature maps accurately detected OOD images. As we showed in our experiment on liver segmentation in MRI, visually identifying OOD data can be quite non-trivial. Therefore, reliable deployment of CNN-based segmentation methods for medical applications requires accurate OOD detection methods to alert the user of the model failure. While this has been a challenging problem because of the massive size and complexity of deep learning models, our proposed method offers an effective solution to this problem.

\bibliographystyle{IEEEtran}
\bibliography{davoodreferences}

\end{document}